\documentclass{article}

\usepackage{microtype}
\usepackage{graphicx}
\usepackage{subfigure}
\usepackage{booktabs} 

\usepackage{amssymb}		
\usepackage{color}			
\usepackage{amsmath}
\usepackage{array}
\usepackage{floatrow}
\newfloatcommand{capbtabbox}{table}[][\FBwidth]
\usepackage{epstopdf}
\usepackage{blindtext}
\usepackage{xcolor}

\usepackage{hyperref}

\usepackage[accepted]{icml2018}
\usepackage{natbib}

\icmltitlerunning{Pseudo-marginal Bayesian inference for supervised Gaussian process latent variable models}

\usepackage{mathtools}
\usepackage{hyperref}
\usepackage{algorithm}
\usepackage{algorithmic}
\usepackage{subfigure}
\usepackage{amsthm}

\newcolumntype{H}{>{\setbox0=\hbox\bgroup}c<{\egroup}@{}}		
\newcommand{\iidsim}{\overset{iid}{\sim}}

\newcommand\sref[1]{Section~\ref{sec:#1}}

\newcommand\fref[1]{Fig.~\ref{fig:#1}}

\newcommand\tref[1]{Table~\ref{tab:#1}}

\newcommand\be{\begin{equation}}
\newcommand\ee{\end{equation}}
\newcommand\bea{\begin{eqnarray*}}
\newcommand\eea{\end{eqnarray*}}

\newcommand\calN{\mathcal{N}}

\newcommand\bx{\mathbf{x}}
\newcommand\bX{\mathbf{X}}
\newcommand\bY{\mathbf{Y}}
\newcommand\by{\mathbf{y}}
\newcommand\bz{\mathbf{z}}

\newcommand\bK{\mathbf{K}}
\newcommand\bu{\mathbf{u}}

\newcommand\bI{\mathbf{I}}

\newcommand\btheta{\boldsymbol{\theta}}

\newcommand\bmu{\boldsymbol{\mu}}

\newcommand\bnu{\boldsymbol{\nu}}
\newcommand\bsigma{\boldsymbol{\sigma}}
\newcommand\bSigma{\boldsymbol{\Sigma}}
\newcommand\bepsilon{\boldsymbol{\epsilon}}

\newcommand{\bbeta}{\boldsymbol{\eta}}

\newcommand{\bZ}{\mathbf{Z}}

\newcommand{\kx}{{k_{x}} }
\newcommand{\ky}{{k_{y}} }
\newcommand{\kz}{{k_{z}} }

\newcommand\Tr{\operatorname{Tr}}

\begin{document}

\twocolumn[
\icmltitle{Pseudo-marginal Bayesian inference for supervised  \\
           Gaussian process latent variable models}


\begin{icmlauthorlist}
\icmlauthor{C. Gadd}{soe,wcpm}
\icmlauthor{S. Wade}{dos}
\icmlauthor{A.A. Shah}{soe}
\icmlauthor{D. Grammatopoulos}{dotm,docb}
\end{icmlauthorlist}

\icmlaffiliation{soe}{School of Engineering, University of Warwick, UK}
\icmlaffiliation{wcpm}{Warwick Center for Predictive Modelling, University of Warwick, UK}
\icmlaffiliation{dos}{Department of Statistics, University of Warwick, UK}
\icmlaffiliation{dotm}{Division of Translational Medicine, Warwick Medical School, UK}
\icmlaffiliation{docb}{Dept of Clinical Biochemistry, Division of Pathology, University Hospital Coventry and Warwickshire, Coventry, UK}

\icmlcorrespondingauthor{C. Gadd}{c.w.l.gadd@gmail.com}

\icmlkeywords{pseudo-marginal, Markov chain Monte Carlo, nonlinear dimension reduction}

\vskip 0.3in
]



\printAffiliationsAndNotice{\icmlEqualContribution} 

\begin{abstract}
	We introduce a Bayesian framework for inference with a supervised version of the Gaussian process latent variable model. The framework overcomes the high correlations between latent variables and hyperparameters by using an unbiased pseudo estimate for the marginal likelihood that approximately integrates over the latent variables. This is used to construct a Markov Chain to explore the posterior of the hyperparameters. 
	We demonstrate the procedure on simulated and real examples, showing its ability to capture uncertainty and multimodality of the hyperparameters and improved uncertainty quantification in predictions when compared with variational inference. 
\end{abstract}

\section{Introduction}
\label{sec:intro}

Statistical Bayesian approaches to regression such as Gaussian process (GP) modelling can be used to learn probabilistic mappings between  inputs and outputs from laboratory or computer experiments based on  {\/\em training\/} samples consisting of 
inputs $\{\bx_n\}_{n=1}^N\subset\mathbb{R}^{\kx}$ and corresponding outputs $\{\by_n\}_{n=1}^N\subset\mathbb{R}^{\ky}$. A GP model can be used to infer a non-linear or latent function by treating the function as a realisation of a (Gaussian) stochastic process. A GP is an infinite collection of random variables such that any finite number have a Gaussian distribution with consistent parameters, and it is fully specified by a mean function and a symmetric positive definite covariance function. In the case of regression, correlations between outputs corresponding to different inputs are specified through the covariance (kernel) function, which, together with a mean function, encapsulates (in a non-parametric manner) any {\/\it a-priori\/} knowledge  and/or assumptions in relation to the target  function, permitting a degree of flexibility in the  assumed complexity and smoothness of the function.


The Gaussian process latent variable model (GPLVM) extends the application of GPs to unsupervised learning tasks for nonlinear dimension reduction. In this case, the inputs are unobserved and treated as latent variables. In \citet{lawrence2005probabilistic}, the latent variables are optimized to the maximum aposteriori solution. To capture uncertainty in the latent variables, \citet{titsias2010bayesian} developed a variational method for GPLVMs. Variational inference in this setting is challenging as the latent variables appear nonlinearly in the inverse of the kernel, making marginalization over these points analytically intractable. This is overcome by assuming a factorized Gaussian variational distribution for the latent variables and through the use of variationally sparse GPs with the augmentation of inducing points. This method yields a tractable lower bound for the log likelihood.


We focus on a supervised version of the GPLVM, which additionally includes a GP prior over latent variables indexed by known and observable inputs. This model was studied in a dynamic setting by \citet{damianou2011variational}. It can further be viewed as a deep GP model \cite{damianou2013deep} with a single hidden layer and observed inputs.


Model hyperparameters defining the covariance function have important implications for smoothness, complexity, and relevance of the inputs. It is common to optimize these parameters based on the approximate marginal likelihood, known as type II maximum likelihood (ML), using gradient-based optimization. However, the likelihood as a function of these parameters is non-convex, and consequently practitioners often find that the optimization is highly dependent on initialisation, with no guarantee of a satisfactory local optimum~\cite{bitzer2010kick}. This is particularly profound when the data set has a low signal-to-noise ratio, which is often the case for medical data. Additionally, while variational methods can  substantially reduce computational time, this comes at the cost of strong assumptions and considerable bias. For example, there are often assumptions of independence, on the forms of distributions, and, dependent upon the choice of divergence, variational methods underestimate or overestimate the variance of these distributions~\cite{blei2017variational}.

Our contribution in this paper is to use this model with a more robust, {\/\it fully Bayesian\/} inference procedure. This allows us to overcome issues with optimization of the hyperparameters and avoids the distributional and independence assumptions of variational methods. Moreover,  we gain an understanding of uncertainty in the hyperparameters and achieve sound quantification of uncertainty in predictions of the response by integrating over the hyperparameters. The natural choice is to explore the posterior distribution of the hyperparameters and latent variables with Markov Chain Monte Carlo (MCMC), but strong correlations between the hyperparameters and latent variables lead to 
low efficiency and poor mixing~\cite{betancourt2015hamiltonian}. We require a method that can break these correlations, which we achieve by using a pseudo-marginal scheme that approximately integrates out the latent variables.


The rest of this paper is structured as follows. In Section~\ref{sec:ProbabilityModel.} we introduce the model before discussing the state-of-the-art variational inference for the model in Section~\ref{sec:Variational.}. A pseudo-marginal Monte Carlo scheme for fully Bayesian inference is then proposed in Section~\ref{sec:PMMC.}. In Section~\ref{sec:ESS}, we describe an elliptical slice sampling scheme for the latent variables, which is necessary to compute the predictions in Section~\ref{sec:Predictions}. 
In Section~\ref{sec:Examples.} we show the advantages of the proposed inference scheme over the variational method in simulated examples and a preliminary case study using medical data with a low signal-to-noise ratio for identifying biomarker associations with antenatal depression. 
Concluding remarks are provided in Section~\ref{sec:Conclusion.}. 

\section{Model}
\label{sec:ProbabilityModel.}

We consider a set of $N$ inputs $\bX:=[\bx_1\hdots\bx_N]^T\in \mathbb{R}^{N\times\kx}$ and corresponding known outputs $\bY:=[\by_1\hdots\by_N]^T\in \mathbb{R}^{N\times\ky}$. This compact matrix notation  is used throughout. Additionally, we introduce a set of unknown latent variable representations $\bZ :=[\bz_1\hdots\bz_N]^T\in \mathbb{R}^{N\times\kz}$ of the outputs, with $\kz \ll \ky$. The $i$-th coordinate (feature) of output $\by_n$ is denoted $y_{n,i}$, where $n$ indexes the data point, and the assumed model is $y_{n,i}=f_i(\bz_n)+\epsilon_{n,i}$, in which 
the noise $\epsilon_{n,i}$ are independent and normally distributed with mean $0$ and precision 
$\beta$, i.e. $\epsilon_{n,i} \iidsim {\cal N} (\epsilon_{n,i}|0,\beta^{-1})$. Independent GP priors (indexed by $\bz$) are placed over the functions $f_i(\bz)$, namely $f_i(\bz)\sim{\cal GP}(0,k_f(\bz,\bz';\btheta))$, where $k_f(\bz,\bz';\btheta)$ is the (common) covariance/kernel function with hyperparameters $\btheta$. 
The notation ${\cal GP}(\cdot,\cdot)$ in which the first argument is the mean function and the second is the covariance function is used throughout. 

Using the notation $f_{n,i}=f_i(\bz_n)$ we can define a matrix ${\bf F}\in\mathbb{R}^{N\times\ky}$ with columns ${\bf f}_{:,i}=(f_{1,i},\hdots,f_{N,i})^T$. A similar notation is used for other matrices, e.g., $\by_{:,i}$ denotes the $i$-th column of $\bY$, i.e., it is the vector with components given by the $i^\text{th}$ feature across all $N$ samples. By the independence assumption 
\begin{equation*}
p({\bf F}|\bZ,\btheta)=\prod_{i=1}^{k_y}p({\bf f}_{:,i}|\bZ,\btheta),
\end{equation*}
and by the properties of GPs, we have $p({\bf f}_{:,i}|\bZ,\btheta)=\mathcal{N} ( {\bf f}_{:,i} | \mathbf{0}, {\bf K}_{f})$, in which ${\bf K}_{f}$ is the $N$ by $N$ 
kernel (covariance) matrix with $n,n'$-th entry $k_f(\bz_n,\bz_{n'};\btheta)$. Thus,
\begin{align}
p(\bY | \bZ,\btheta, \beta) &= \int \prod_{i=1}^{\ky}\prod_{n=1}^{N}p(y_{n,i} | f_{n,i},\beta) p( {\bf f}_{:,i}  | \bZ, \btheta ) d{\bf F}  \nonumber \\
&= \prod_{i=1}^{\ky} \mathcal{N}( \by_{:,i}|\mathbf{0},{\bf K}_{f} + \beta^{-1} \mathbf{I}_N),
\label{eq:PriorY}
\end{align}
in which $p(\mathbf{y}_{:,i}|{\bf f}_{:,i},\beta) = \mathcal{N}(\by_{:,i} | {\bf f}_{:,i},\beta^{-1}\mathbf{I}_N)$ by virtue of the noise model. 
The model for the latent points comes in the form of independent GP priors $z_j(\bx)\sim {\cal GP} (0,k_z(\bx,\bx';\bsigma))$, $j=1,\hdots,\kz$. Thus,
\begin{equation}
\label{eq:CorrelatedPriorZ.}
\begin{split}
p(\bZ | \bX, \bsigma) &= \prod_{j=1}^{\kz}p(\bz_{:,j} | \bX, \bsigma)=\prod_{j=1}^{\kz}\calN(\bz_{:,j} | \mathbf{0}, {\bf K}_{z}),
\end{split}
\end{equation}
where 
${\bf K}_{z}$ is the $N$ by $N$ the kernel matrix, with $n,n'$-th entry equal to $k_z(\bx_n,\bx_{n'};\bsigma)$. The joint density for the observed data and latent variables is:
\begin{equation*}
\label{ProbabilityModel.}
p({\bf Y},{\bf Z}| \bX,\btheta,\bsigma,\beta) = p(\bY|\bZ,\btheta,\beta)p(\bZ|\bX,\bsigma).
\end{equation*} 
The Bayesian model is completed with a prior on the precision $\beta$ and kernel hyperparameters $\btheta$ and $\bsigma$, which we assume factorizes as
$$ p(\beta,\btheta,\bsigma)=p(\beta)p(\btheta)p(\bsigma).$$


\section{Variational Bayes}
\label{sec:Variational.}
%
%

State-of-the-art inference for GPLVMs is based on a variational Bayes method \cite{titsias2010bayesian}, which assumes a factorized Gaussian variational posterior for the latent variables.  Ideas from variationally sparse GPs with auxiliary inducing variables are utilized to derive a tractable lower bound to marginal likelihood, which can then be optimized with respect to hyperparameters and variational parameters. We briefly describe an application of this variational method for the supervised GPLVM presented in the previous section, which is similar to the variational approach of \citet{damianou2011variational} but extended for multivariate inputs and the variational approach of \citet{damianou2015deep} but for a single hidden layer and observed inputs.


A variational distribution $q(\bZ)$ for the latent variables (approximating the true posterior $p(\bZ | \bX,\bY, \btheta, \bsigma,\beta)$) is taken to have the factorized Gaussian form:
\begin{equation*}
\label{eq:variationaldistribution}
q(\bZ)= \prod_{j=1}^{\kz}\calN(\bz_{:,j} | \bmu_j,{\bf S}_j),
\end{equation*}
where ${\bf S}_j$ is a full $N$ by $N$ covariance matrix. 
Using Jensen's inequality, we can derive the evidence lower bound: 
\begin{align}
&\log p(\bY|\bX,\btheta,\bsigma,\beta) \nonumber \\
& \geq \int q(\bZ) \log \left[ \frac{p(\bY|\bZ,\btheta,\beta)p(\bZ | \bX,\bsigma)}{q(\bZ)}\right]d\mathbf{\bZ} \nonumber \\
&= \mathcal{\tilde{F}}(q(\bZ),\btheta,\beta) - \mathbf{\text{KL}}\left(q(\bZ)||p(\bZ | \bX,\bsigma)\right). \label{eq:F}
\end{align}
%
The second term is the negative KL divergence 
between $q(\bZ)$ and  $p(\bZ | \bX,\bsigma)$, which can be analytically computed since both are Gaussians. 
Given  that the data $\{\by_n\}_{n=1}^N$ is conditionally independent across features, we can express the first term $\mathcal{\tilde{F}}(q(\bZ),\btheta,\beta)$ in (\ref{eq:F}) as:
\begin{align}
\mathcal{\tilde{F}}(q(\bZ),\btheta,\beta) &= \sum_{i=1}^{\ky}\int q(\bZ) \log p\left(\by_{:,i} | \bZ,\btheta,\beta\right)d\bZ \nonumber \\ &:=\sum_{i=1}^{\ky}\mathcal{\tilde{F}}_{i} (q\left(\bZ\right),\btheta,\beta). \label{eq:Ftilde}
\end{align}
In the current form for $\mathcal{\tilde{F}}\left(q\left(\bZ\right),\btheta,\beta\right)$ in Eq.~\eqref{eq:Ftilde}, $\bZ$ remains inside the inverse of the kernel, making expectations under distributions of $\bZ$ difficult to compute. In order to formulate a tractable problem, we employ the variational sparse GP approach of~\citet{titsias2009variational}. For each vector of latent function values ${\bf f}_{:,i}$, $i=1,\hdots,\ky$, we introduce a separate set of $M$ auxiliary {\/\em inducing function values\/} $\mathbf{u}_{:,i}\in~\mathbb{R}^M$, evaluated at a set of $M$ inducing point locations  
given by the matrix $\bZ_{\bu}\in\mathbb{R}^{M\times \kz}$. 
The augmented probability model is:
\begin{equation*}
\begin{split}
&p(\by_{:,i}, {\bf f}_{:,i}, \bu_{:,i} | \bZ,\btheta,\beta, \bZ_{\bu}) \\
&=p(\by_{:,i} | {\bf f}_{:,i},\beta)p({\bf f}_{:,i} | \bZ,\btheta,\bu_{:,i},\bZ_{\bu})p(\bu_{:,i} | \bZ_{\bu},\btheta),
\end{split}
\end{equation*}
and from the joint GP prior over  $({\bf f}_{:,i},\bu_{:,i})$,
\begin{equation*}
\begin{split}
\label{eq:conditionalGPprior1}
&p({\bf f}_{:,i} | \bZ,\btheta, \bu_{:,i},\bZ_{\bu})  \\
&= \mathcal{N}({\bf f}_{:,i} | \pmb\alpha_i, {\bf K}_{f}-{\bf K}_{fu}{\bf K}_{ u}^{-1} {\bf K}_{uf}),
\end{split}
\end{equation*}
and $p(\bu_{:,i} | \bZ_{\bu},\btheta) = \mathcal{N}(\bu_{:,i} | \mathbf{0},{\bf K}_{u})$, in which ${\bf K}_{ u}$ is the covariance matrix corresponding to the inducing points, ${\bf K}_{fu}={\bf K}_{ uf}^T$ is the cross-covariance between the inducing and latent points and $\pmb\alpha_i := {\bf K}_{fu}{\bf K}_{u}^{-1}\bu_{:,i}$.
%
%
We then use a variational approach again to approximate the true posterior of $({\bf f}_{:,i}, \bu_{:,i})$ with the variational distribution: 
\begin{equation*}
q({\bf f}_{:,i},\bu_{:,i})=p({\bf f}_{:,i} | \bZ,\btheta, \bu_{:,i},\bZ_{\bu})\phi(\bu_{:,i}).
\end{equation*}
%
Lower bounds on the log likelihood term in the integrand of $\mathcal{\tilde{F}}_{i}$ in Eq.~\eqref{eq:Ftilde} are given by: 
\begin{align}
&\log p(\by_{:,i}|\bZ,\btheta,\beta) \nonumber \\
&\geq \int \phi(\bu_{:,i})  \log \frac{p(\bu_{:,i}| \bZ_{\bu},\btheta)\mathcal{N}(\by_{:,i}|\pmb\alpha_i,\beta^{-1}I_N)}{\phi(\bu_{:,i})}d\bu_{:,i} \nonumber \\
&\quad- \frac{\beta}{2}\Tr ({\bf K}_{f}-{\bf K}_{fu}{\bf K}_{u}^{-1}{\bf K}_{ uf}). \label{eq:LB1}
\end{align}
%
Combining Eq.~\eqref{eq:Ftilde} with Eq.~\eqref{eq:LB1} and interchanging the order of integration (under the assumption that $\phi(\bu_{:,i})$ does not depend on $\bZ$), we have
%
%
\begin{equation*}
\begin{split}
&\mathcal{\tilde{F}}_{i} (q(\bZ),\btheta,\beta,\bZ_u) \geq \int \phi(\bu_{:,i}) \log \frac{p(\bu_{:,i}| \bZ_{\bu},\btheta)}{\phi(\bu_{:,i})} d\bu_{:,i} \\
&+ \int \phi(\bu_{:,i}) \langle\log \mathcal{N}(\by_{:,i}|\pmb\alpha_i, \beta^{-1}\mathbf{I}_N)\rangle_{q}d\bu_{:,i} \\
&- \frac{\beta}{2}\Tr (\langle{\bf K}_{f}\rangle_{q})+\frac{\beta}{2}\Tr ({\bf K}_{u}^{-1}\langle{\bf K}_{uf}{\bf K}_{fu}\rangle_{q}).
\end{split}
\end{equation*}
in which $\langle\rho\rangle_{q}$ 
denotes the expected value of a quantity $\rho$ under the distribution $q(\bZ)$. 
The optimal setting of $\phi(\bu_{:,i})$ by maximizing the lower bound is
\begin{equation*}
\phi(\bu_{:,i})\propto \exp \left(\langle\log \mathcal{N}(\by_{:,i} | \pmb\alpha_i,\beta^{-1}\mathbf{I}_N)\rangle_{q} \right)p(\bu_{:,i}| \bZ_{\bu},\btheta),
\end{equation*}
and the lower bound that incorporates such an optimal setting is obtained as follows:
\begin{equation*}
\label{eq:Ftilded}
\begin{split}
&\mathcal{\tilde{F}}_{i} \left(q\left(\bZ\right), \btheta, \beta,\bZ_u \right) \\
&\geq  \log \bigg(\int e^{\langle\log \mathcal{N}\left(\by_{:,i} | \pmb\alpha_i,\beta^{-1}\mathbf{I}_N\right)\rangle_{q}}  p(\bu_{:,i}| \bZ_{\bu},\btheta)d\bu_{:,i}\bigg) \\
&- \frac{\beta}{2}\Tr (\langle{\bf K}_{ f}\rangle_{q}) + \frac{\beta}{2}\Tr ({\bf K}_{u}^{-1}\langle{\bf K}_{uf}{\bf K}_{fu}\rangle_{q}).
\end{split}
\end{equation*}
For a number of kernels this lower bound can be computed in closed form. In the case of a squared exponential kernel with automatic relevance determination, this amounts to computing the statistics~\cite{titsias2010bayesian}: 

\begin{equation*}
\begin{split}
\psi_0 &= \langle \Tr  \left({\bf K}_f\right)\rangle_{q}, \quad \psi_1 = \langle {\bf K}_{ fu}\rangle_{q},\quad\psi_2 = \langle {\bf K}_{uf}{\bf K}_{ fu} \rangle_{q}.
\end{split}
\end{equation*}

We can now optimize the tractable variational lower bound of the supervised GPLVM with respect to the variational parameters $(\left\{\bmu_j,{\bf S}_j\right\}_{j=1}^\kz,\bZ_{\bu})$ and hyperparameters $(\btheta, \bsigma, \beta)$. We note that when the hyperparameters are fixed (as is the case in following section), the variational parameters are optimized by using scaled conjugate gradients (SCG) with the gradient derivations detailed in~\citet{damianou2015deep}.  
The training procedure with fixed hyperparameters is given in Algorithm~\ref{Alg:LVM}. 

\begin{algorithm}
\caption{Variational Bayes with fixed  $(\btheta, \bsigma, \beta)$.}\label{Alg:LVM}
	\begin{algorithmic}[1]
		\REQUIRE Data set $\left(\bX, \bY\right)$.
		\newline
		\STATE Initialise: Variational distribution means using principal component analysis on $\bY$ and variances to neutral values, $0.5$.\\
		
		\WHILE{not at \textit{local} maximum of variational lower bound}
		\STATE Compute the $\psi$ statistics analytically.
		\STATE Use to compute the variational lower bound.
		\STATE Adjust variational parameters $(\left\{\bmu_j,{\bf S}_j\right\}_{j=1}^\kz,\bZ_{\bu})$ 
		according to gradients.
		\ENDWHILE
	\end{algorithmic}
\end{algorithm}

\section{Bayesian inference}
\label{sec:PMMC.}

\subsection{Collapsed Gibbs through the pseudo-marginal}

In our fully Bayesian setting, we aim to infer the posterior distribution for all hyperparameters and latent variables and integrate with respect to their posterior distributions when making predictions to achieve sound uncertainty quantification. As opposed to the the variational method described in the previous section, this overcomes issues associated to optimization of the hyperparameters and avoids the distributional and independence assumptions, in addition to improving uncertainty quantification in prediction. 

In the supervised GPLVM model, the latent parameters and hyperparameters are strongly coupled, leading to sharp peaks in their posterior when latent variables are fixed. In a Gibbs sampling algorithm which alternates between sampling and fixing the latent variables and hyperparameters, this leads to poor MCMC mixing and slow convergence rates~\cite{filippone2014pseudo}. 
Although analytical integration of the latent variables $\bZ$ is intractable since they appear nonlinearly in the inverse kernel matrix ${\bf K}_{f}$, we can break the correlation between the latent variables and hyperparameters by approximately integrating over the latent variables through a pseudo-marginal Monte Carlo scheme.
The results of~\citet{andrieu2009pseudo} and~\citet{beaumont2003estimation} reveal that we can use an unbiased estimate of the marginal  likelihood to sample from the correct hyperparameter posterior distribution. 

Specifically, we use importance sampling to obtain an unbiased approximation to the marginal likelihood based on an approximate distribution $q\left(\bZ\right) \approx p\left(\bZ|\mathcal{D},\bsigma,\btheta,\beta \right)$.  Drawing $Q$ importance samples, the unbiased estimate of the marginal is:

\begin{equation}
\label{eq:approx_marginal}
\begin{split}
\tilde{p}(\bY|\bX,\bsigma,\btheta,\beta) &\simeq \frac{1}{Q}\sum_{q=1}^{Q}\frac{p(\bY|\bZ^{(q)},\btheta,\beta)p(\bZ^{(q)} \mid\bX,\bsigma)}{q(\bZ^{(q)})},
\end{split}
\end{equation}

where $\bZ^{(q)} \iidsim q(\bZ)$, and  $p(\bY|\bZ^{(q)},\btheta,\beta)$ and $p(\bZ^{(q)}|\bX,\bsigma)$ are the GP models given by Eqs.~\eqref{eq:PriorY} and~\eqref{eq:CorrelatedPriorZ.} respectively. We select $q(\bZ)$ to be the approximate variational posterior. This pseudo-marginal can be used to sample from the posterior of the hyperparameters in a Metropolis-Hastings algorithm.

To improve mixing, we split our set of hyperparameters $\pmb{\xi}=\left(\bsigma,\btheta,\beta\right)$ into $R$ adjoint subsets, and update each block, denoted by $\pmb{\xi}_r$, $r=1,\ldots,R$, in a Metropolis-Hastings within Gibbs algorithm. Each block $\pmb{\xi}_r$ is updated with a random walk based on a transformation $\bbeta_r = t_r(\pmb{\xi}_r)$ to ensure full support on the real space of appropriate dimension. We use a multivariate normal proposal distribution for the transformed parameter: $\pi(\bbeta_r'|\bbeta_r)\sim\mathcal{N}(\mathbf{0},\bSigma_r)$, which gives the proposal distribution 
$ \pi(\pmb{\xi}_r) = |\partial t_r/ \partial  \pmb{\xi}_r| \pi(\bbeta_r)$ in the original parameter space. The acceptance probability for a move from $\pmb{\xi}_r$ to $\pmb{\xi}'_r$ is therefore:

\begin{equation*}
\label{eq:acceptanceProbability.}
\tilde{\alpha}(\pmb{\xi}_r,\pmb{\xi}'_r) = \text{min}\left[1,  \frac{\tilde{p}(\bY|\bX,\pmb{\xi}') p(\pmb{\xi}'_r)}{ \tilde{p}(\bY|\bX,\pmb{\xi}) p(\pmb{\xi}_r)}\frac{|\partial t_r/ \partial  \pmb{\xi}_r (\pmb{\xi}_r)|}{|\partial t_r/ \partial  \pmb{\xi}_r (\pmb{\xi}_r')|}\right],
\end{equation*} 
%

%
We now employ a variant of the adaptive Metropolis-Hastings algorithm in~\citet{haario2001adaptive}, which adapts the proposal covariance matrix $ \bSigma_r$ to approximate the target distribution's covariance matrix multiplied by a constant $s_d$. Following~\citet{haario2001adaptive}, we choose $s_d=2.38^2 / d$ where $d$ is the dimension of the block. We begin by choosing an initial proposal covariance matrix for each block, and after $g_0$ iterations, we adaptively update the covariance matrix using the sample covariance, with a small positive constant on the diagonal.
%
%
The full procedure is outlined in Algorithm~\ref{Alg:AMAlgorithm.}.

\begin{algorithm}[H]
	\caption{Pseudo-marginal adaptive MH in Gibbs.}\label{Alg:AMAlgorithm.}
	\begin{algorithmic}[1]
		\small
		\FOR{$g=1,2,\dots$} 
		\FOR{each $\pmb{\xi}_r$, $r=1,\ldots,R$ }
		
		\STATE Sample $\bbeta_{r}'=\bbeta_{r}^{(g)}+\bepsilon_g$ where $\bepsilon_g\sim \mathcal{N}(\mathbf{0},\bSigma_r^{(g-1)})$.
		\STATE Find the unbiased approximation $\tilde{p}(\bY|\bX,\pmb{\xi}')$ using importance sampling (Eq.~\eqref{eq:approx_marginal}).
		\STATE Set:
		\begin{equation*}
		\pmb{\xi}_r^{(g)} = \left\{ 
		\begin{array}{ll}
		\pmb{\xi}_r' \;\;\; \text{ with probability } \tilde{\alpha}(\pmb{\xi}_r^{(g-1)},\pmb{\xi}_r') \\
		\pmb{\xi}_r^{(g-1)} \text{ with probability } 1- \tilde{\alpha}(\pmb{\xi}_r^{(g-1)},\pmb{\xi}_r')
		\end{array}
		\right. .
		\end{equation*}
		\IF{$g>g_0$}
		\STATE $\bSigma_r^{\left(g\right)} =\frac{s_d}{g-1}\left[\sum\limits_{m=1}^{g}\bbeta_r^{\left(m\right)}\bbeta_r^{(m)^T}-g\bar{\bbeta_r}\bar{\bbeta_r}^T  \right] + s_d \epsilon \bI $.
		\ENDIF
		\ENDFOR
		\STATE \textbf{return} $\pmb{\xi}^{(g)}$ for $g> n_0$.
		\ENDFOR
	\end{algorithmic}
\end{algorithm}

\subsection{Uncollapsing with elliptical slice sampling}
\label{sec:ESS}


We sample the latent variables using an elliptical slice sampler conditioned on the hyperparameters sampled in the pseudo-marginal scheme. These samples will be used to compute predictions in Section \ref{sec:Predictions}. The target distribution for the sampler is the full conditional of the latent variables:
\begin{equation*}
\begin{split}
&p\left(\bZ|\bY,\bX,\pmb{\xi}\right)	\propto p\left(\bY|\bZ,\btheta,\beta\right) p\left(\bZ|\bsigma,\bX \right) \\
&\propto  \prod_{i=1}^{\ky}\mathcal{N}\left(\by_{:,i}|\mathbf{0},\bK_f\left(\bZ,\bZ;\btheta\right)+\beta^{-1}\mathbf{I}_N
\right) \times \\
&\quad  \prod_{j=1}^{\kz}\mathcal{N}\left(\bz_{:,j} | 0,\bK_z\left(\bX,\bX; \bsigma\right)\right).
\end{split}
\end{equation*}
%
%
We use the elliptical slice sampling algorithm of ~\citet{murray2010elliptical}, in which a new state $\bZ'$ is proposed by
%
%
\begin{equation*}
\bZ' =  \bnu \sin \alpha + \bZ \cos \alpha, \qquad 	\bnu_{:,j}\iidsim\mathcal{N}\left(0,\bK_z\right).
\end{equation*}
This defines a full ellipse passing through our previous state $\bZ$ and a prior sample $\bnu\in\mathbb{R}^{N\times\kz}$ as $\alpha$ varies. This proposal depends on a tuning parameter $\alpha$ which would be chosen \textit{a priori} under a normal Metropolis Hastings scheme. The algorithm of \citet{murray2010elliptical} adaptively choses this tuning parameter using slice sampling. 
The procedure for sampling $\bZ$ using the elliptical slice sampler is given in Algorithm~\ref{Alg:ESSAlgorithm.}. 

\section{Predictions using Monte Carlo}
\label{sec:Predictions}

The marginalised predictive density for a test point $\bx_*$ is:
\begin{align}
&p\left(\by_*|\bx_*,\bY,\bX \right) =\int p\left(\by_*|\bz_*, \bZ, \bY, \btheta, \beta \right) \times\nonumber \\
 &\quad  p\left(\bz_*| \bx_*, \bX,\bZ,\bsigma \right) p\left( \bZ, \pmb{\xi} | \bY, \bX \right) d \bz_*  d\bZ d\pmb{\xi}. \label{eqn:MonteCarloPredictive.}
\end{align}
The second term inside the integral of Eq.~\eqref{eqn:MonteCarloPredictive.} is the predictive density of the latent variable $\bz_*$ given the latent variables, hyperparameters, and data, which is given by the GP predictive density:
\begin{align}
\label{eq:PredZ}
p\left(\bz_* | \bx_*, \bX, \bZ,\bsigma\right) =\prod_{j=1}^\kz \mathcal{N}\left(z_{*j}| \bK_{z*}^T\bK_z^{-1} \bz_{:,j}, s_{*}\right),
\end{align}
with $s_{*}=k_z( \bx_*, \bx_*;\bsigma) - \bK_{z*}^T\bK_z^{-1} \bK_{z*}$. Here 
$\bK_{z*}$ is the cross-covariance at the training inputs $\bX$ and the test input $\bx_*$. Similarly, the first term  inside the integral of Eq.~\eqref{eqn:MonteCarloPredictive.} is the predictive density of the test output $\by_*$ given $\bz_*$, the latent variables, hyperparameters, and data, which is given by the GP predictive density:
\begin{equation*}
\begin{split}
p\left(\by_* | \bz_*, \bZ, \bY,\btheta,\beta\right) = \prod_{i=1}^\ky \mathcal{N}\left(y_{*i} | \mathbf{A}\by_{:,i},\mathbf{S}+\beta^{-1}\right),
\end{split}
\end{equation*}

\begin{algorithm}
	\caption{Elliptical slice sampler for the latent variables.}
	\label{Alg:ESSAlgorithm.}
	\begin{algorithmic}[1]
		\small
		\REQUIRE current state $\bZ$, and log-likelihood function.
		\ENSURE new state $\bZ'$.
		\newline
		\STATE Sample: $\bnu\sim\prod_{j=1}^{\kz}\mathcal{N}\left(\bnu_{:,j}|0,\bK_z\right)$, creating an ellipse at current state with  $\alpha=0$.
		\STATE Log-likelihood threshold:
		\begin{align*}
		u \sim \text{Uniform}\left[0,1\right], \quad \log h \leftarrow\log  p\left(\bY|\bZ;\btheta,\beta\right) + \log u.
		\end{align*}
		\STATE Draw an initial proposal, define bracket on the ellipse:
		\begin{align*}
		\alpha\sim \text{Uniform}\left[0,2\pi\right], \quad \left[\alpha_\text{min},\alpha_\text{max}\right]\leftarrow[\alpha-2\pi,\alpha].
		\end{align*}
		\WHILE{not returned}
		\STATE Propose new latent variables: 
		\begin{equation*}
		\bZ'\leftarrow \bnu \sin \alpha + \bZ\cos \alpha. 
		\end{equation*}
		\STATE {\bf if:} $\log p\left(\bY|\bZ',\btheta,\beta\right)  > \log h$ (proposal lies in slice) {\bf then:}
		\STATE \hspace{10mm} Accept: {\bf return $\bZ'$}.
		\STATE {\bf else:}
		\STATE \hspace{10mm} Shrink bracket and re-sample point from ellipse:
		\STATE \hspace{10mm} {\bf if} $\alpha < 0$ {\bf then:} $\alpha_\text{min} \leftarrow \alpha$ {\bf else:} $\alpha_\text{max} \leftarrow \alpha$
		\STATE \hspace{10mm} $\alpha \sim \text{Uniform}\left[\alpha_\text{min},\alpha_\text{max}\right]$.
		\ENDWHILE
	\end{algorithmic}
\end{algorithm}

where $\mathbf{A}=\bK_{f*}^T(\bK_f + \beta^{-1}\mathbf{I}_N)^{-1}$ and $\mathbf{S}=\bK_{f**} - \bK_{f*}^T(\bK_f + \beta^{-1}\mathbf{I}_N)^{-1} \bK_{f*}$. Here $\bK_{f**}$ corresponds to the covariance at $\bz_*$ and $\bK_{f*}$ corresponds to cross-covariance at $\bZ$ and $\bz_*$.

Our MCMC samples can be used to obtain an approximation to the marginalised predictive density in Eq.~\eqref{eqn:MonteCarloPredictive.}. However, the latent variable $\bz_*$ cannot be marginalized analytically. Thus, given each sample of the chain $(\pmb{\xi}^{(g)}, \bZ^{(g)})$, we sample the latent variable $\bz_*^{(g)}$ based on its predictive distribution in Eq.~\eqref{eq:PredZ}. The predictive density estimate is
\begin{equation*}
\label{eqn:marginalisedPredictive}
\begin{split}
&p\left(\by_*|\bx_*,\bY,\bX \right) \approx \frac{1}{G} \sum_{g=1}^G p\left(\by_* | \bz_*^{(g)}, \bZ^{(g)}, \bY,\btheta^{(g)},\beta^{(g)}\right).
\end{split}
\end{equation*}

Similarly, we can estimate the posterior mean function with

\begin{equation*}
\label{eqn:marginalisedPredictiveMean}
\begin{split}
&\mathbb{E}\left[\by_*|\bx_*,\bY,\bX \right] \approx \frac{1}{G} \sum_{g=1}^G \bK_{f*}^{(g)\,T}(\bK_f^{(g)} + \beta^{(g)\,-1}\mathbf{I}_N)\bY.
\end{split}
\end{equation*}

\section{Examples}
\label{sec:Examples.}

\subsection{Simulated: sinusoidal data}
\label{sec:ToyProblem.}

\begin{figure*}[h]
	\centering
	\subfigure[Case 1: $\sigma_1$ v. $\beta^{-1}$.]{
		\label{fig:Toy_well_specified_Joint1}
		\includegraphics[width=30mm]{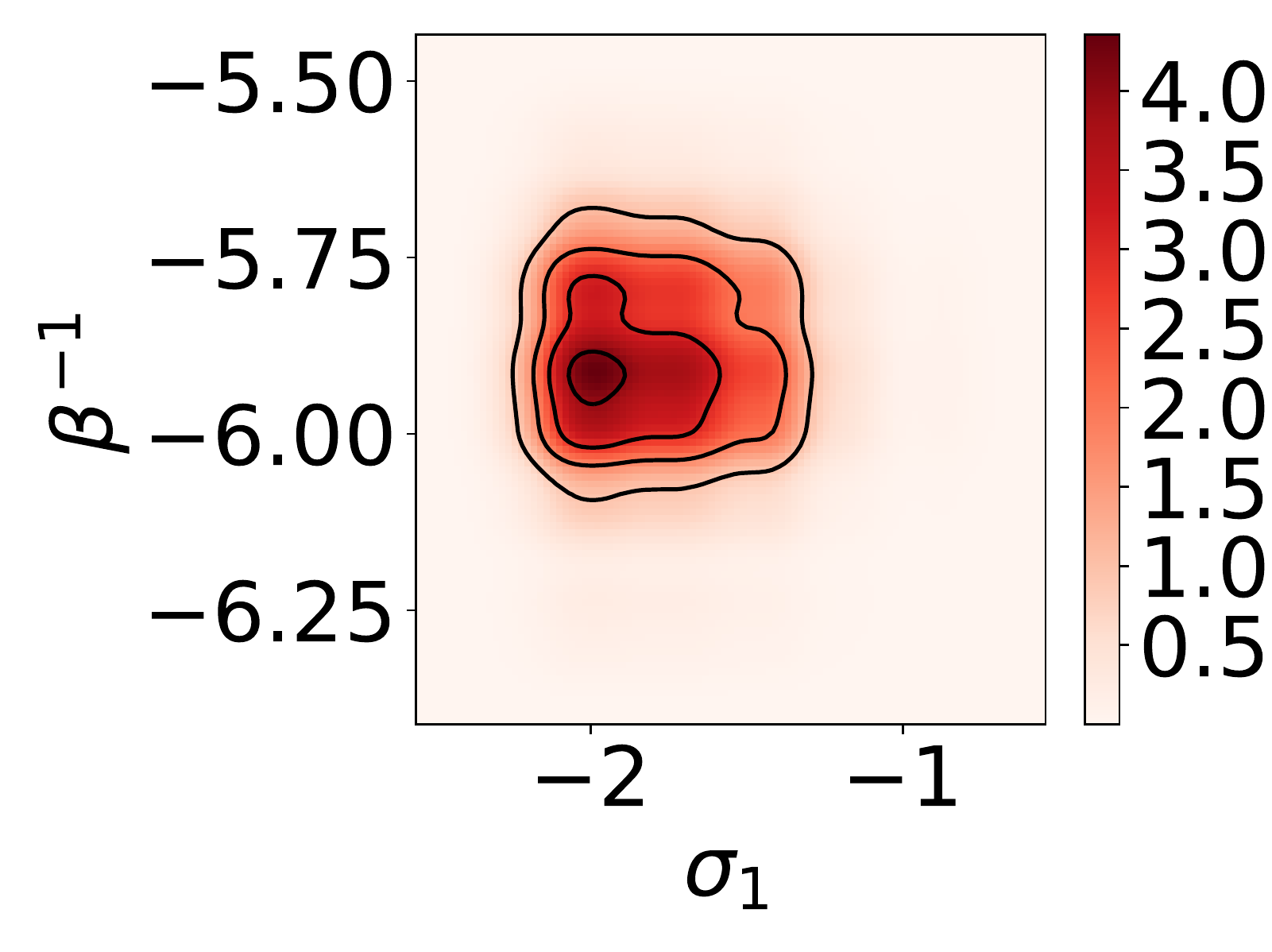}
	}
	\subfigure[Case 1: $\theta_1$ v. $\theta_2$.]{
		\label{fig:Toy_well_specified_Joint2}
		\includegraphics[width=30mm]{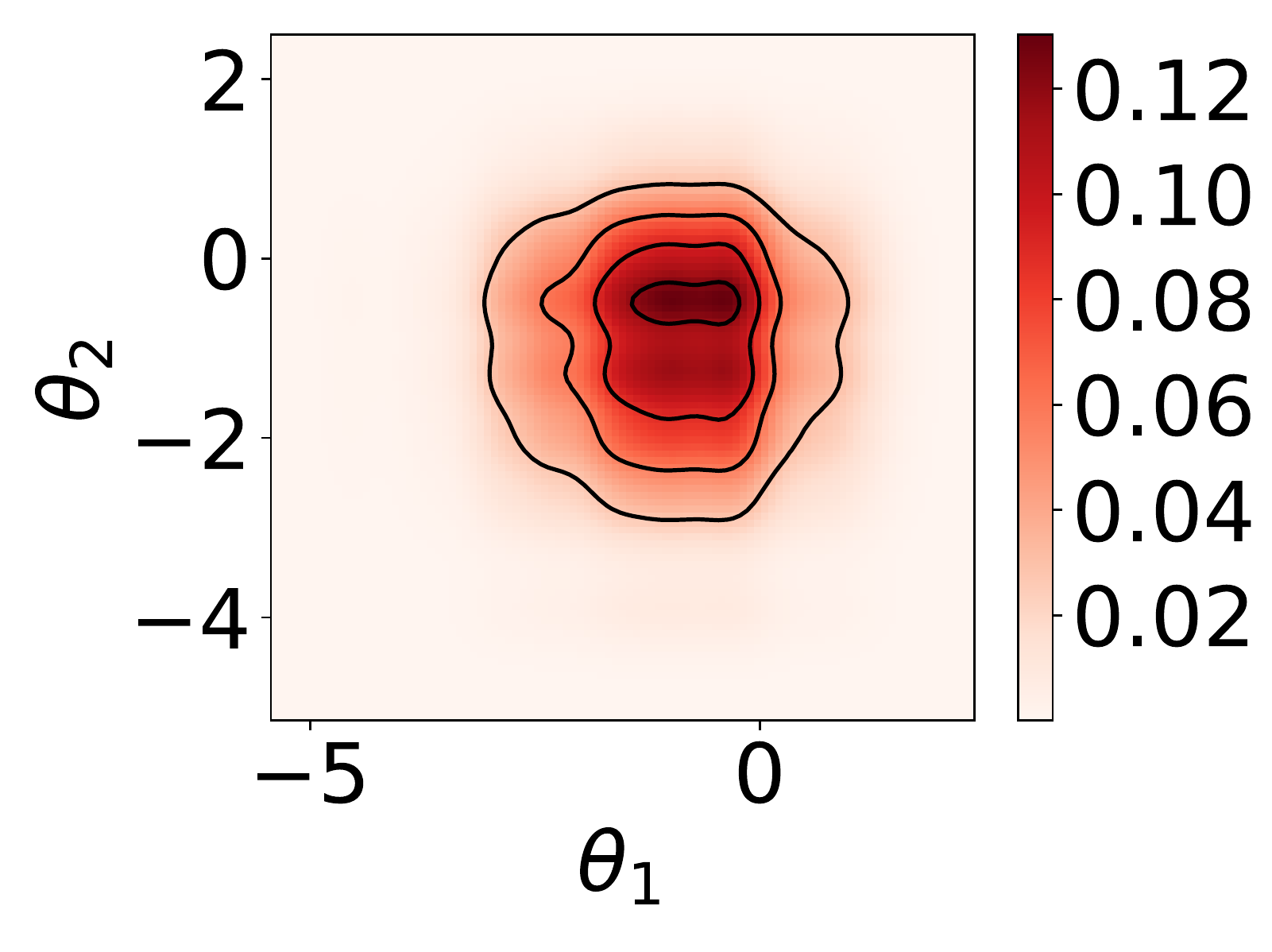}
	} 
	\subfigure[Case 1: $\theta_S$ v. $\beta^{-1}$.]{
		\label{fig:Toy_well_specified_Joint3}
		\includegraphics[width=30mm]{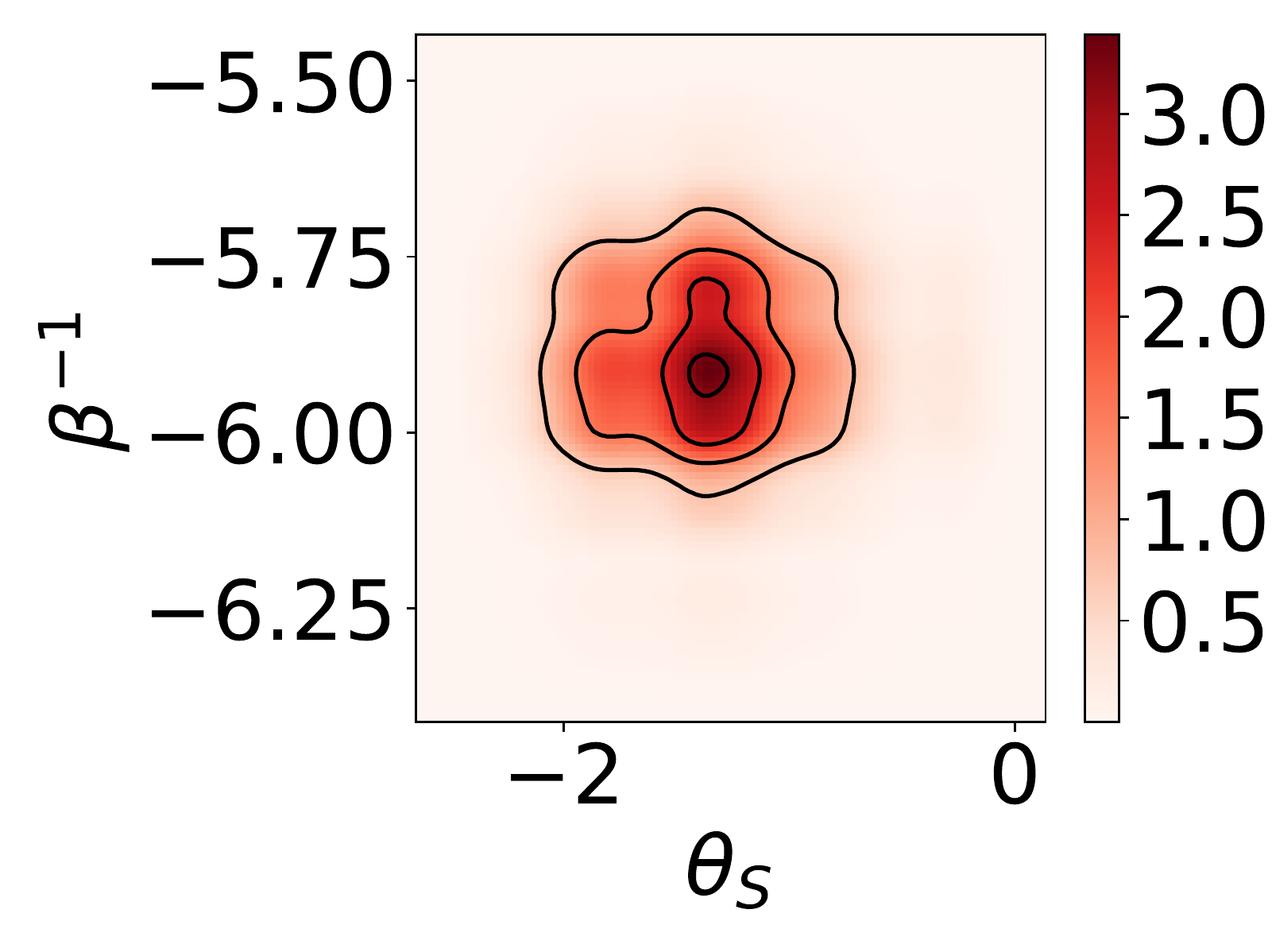}
	}
\\
	\subfigure[Case 2: $\sigma_1$ v. $\beta^{-1}$]{
		\label{fig:Toy_mix1_Joint1}
		\includegraphics[width=30mm]{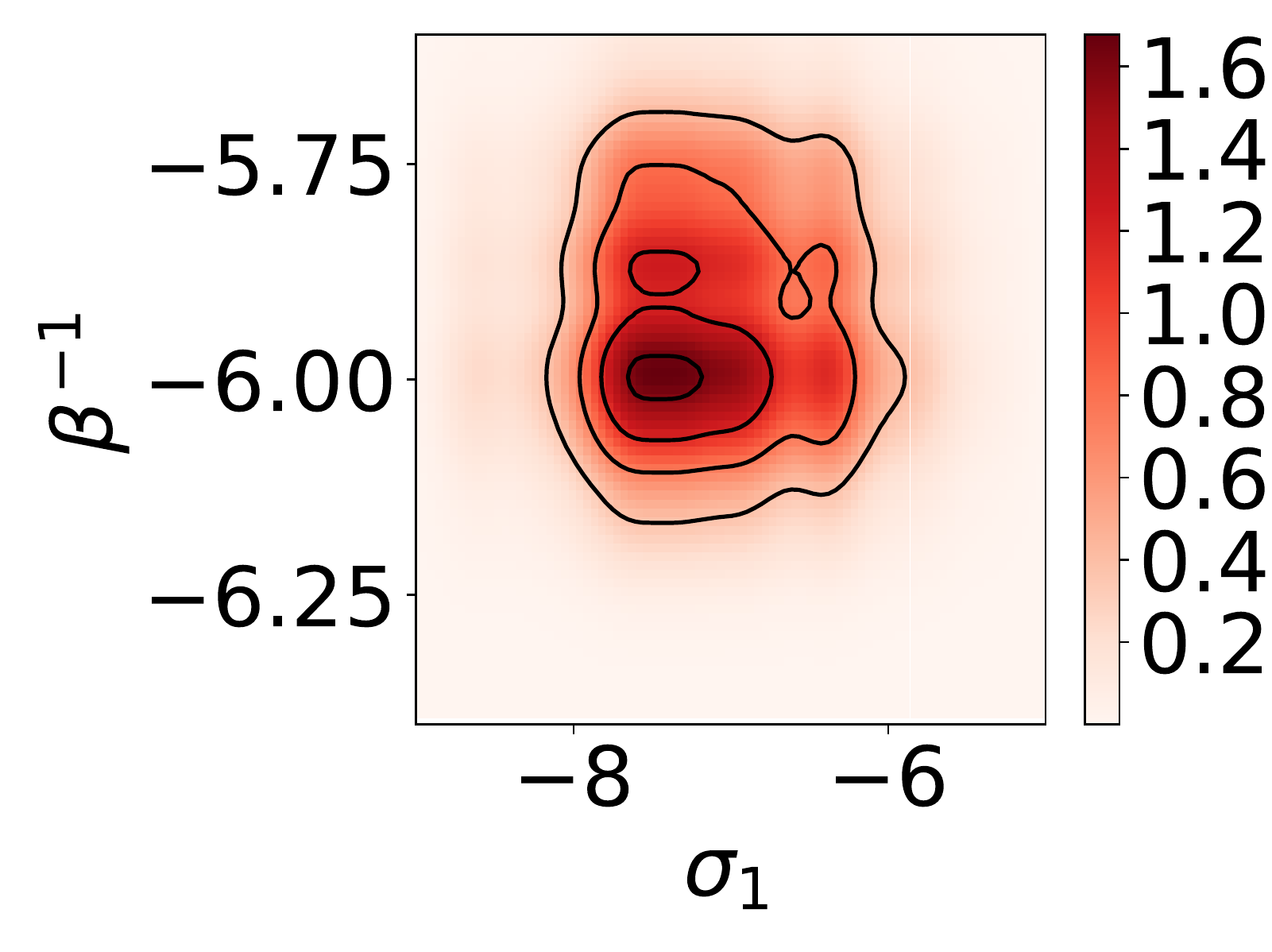}
	}
	\subfigure[Case 2: $\theta_1$ v. $\theta_2$]{
		\label{fig:Toy_mix1_Joint2}
		\includegraphics[width=30mm]{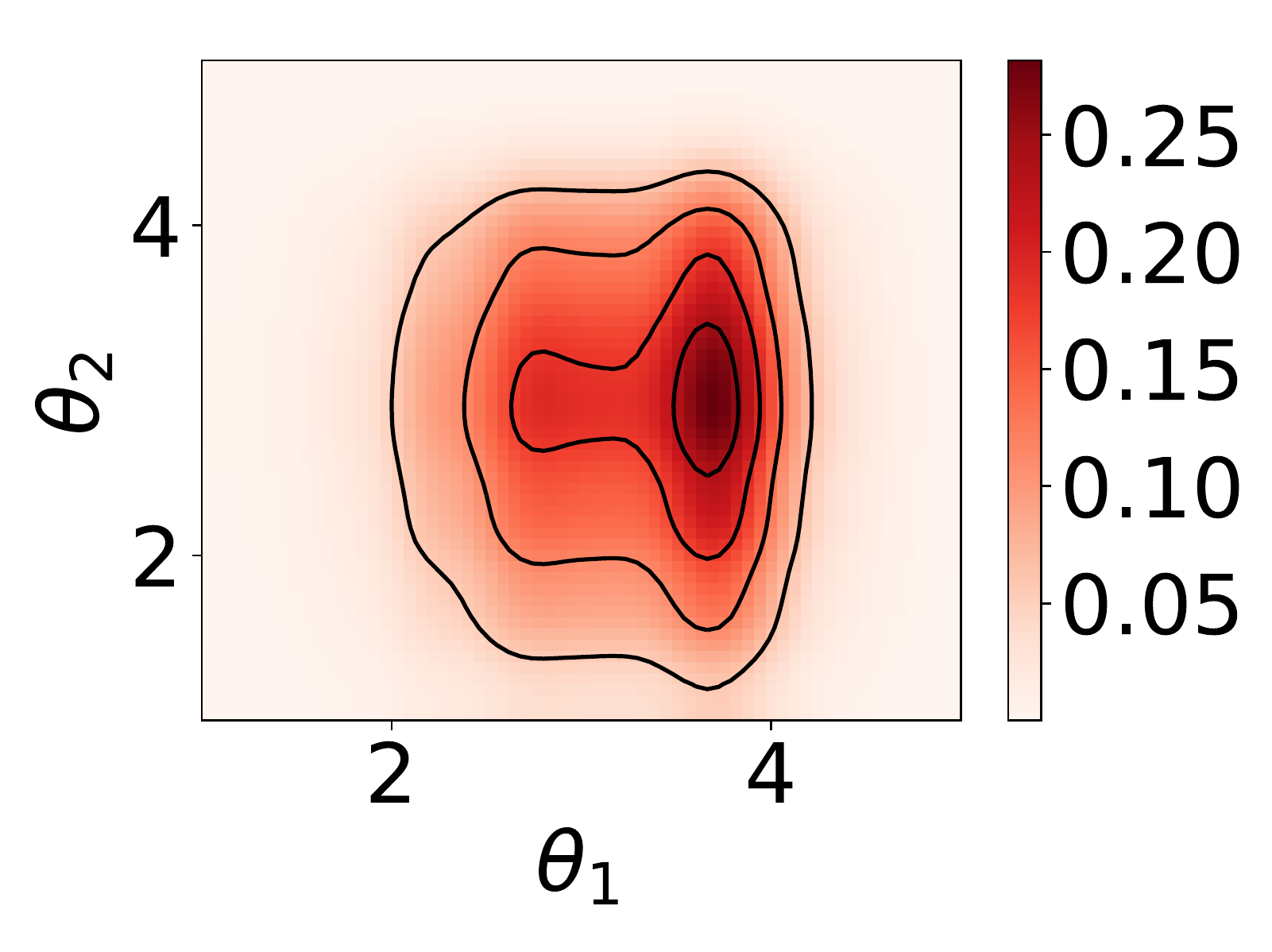}
	} 
	\subfigure[Case 2: $\theta_S$ v. $\beta^{-1}$]{
		\label{fig:Toy_mix1_Joint3}
		\includegraphics[width=30mm]{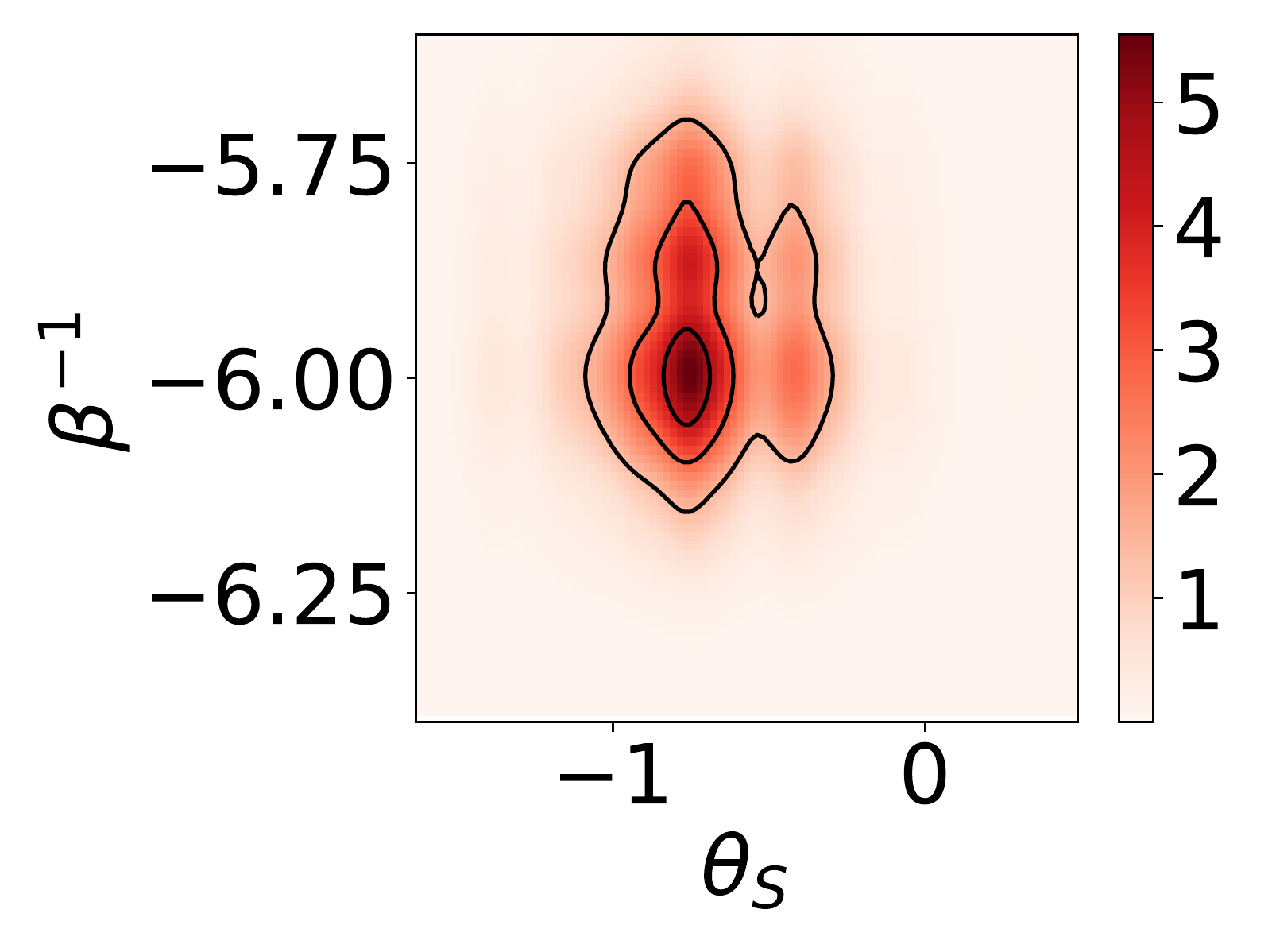}
	}
\\
	\subfigure[Case 3: $\sigma_1$ v. $\beta^{-1}$]{
		\label{fig:Toy_mix2_Joint1}
		\includegraphics[width=30mm]{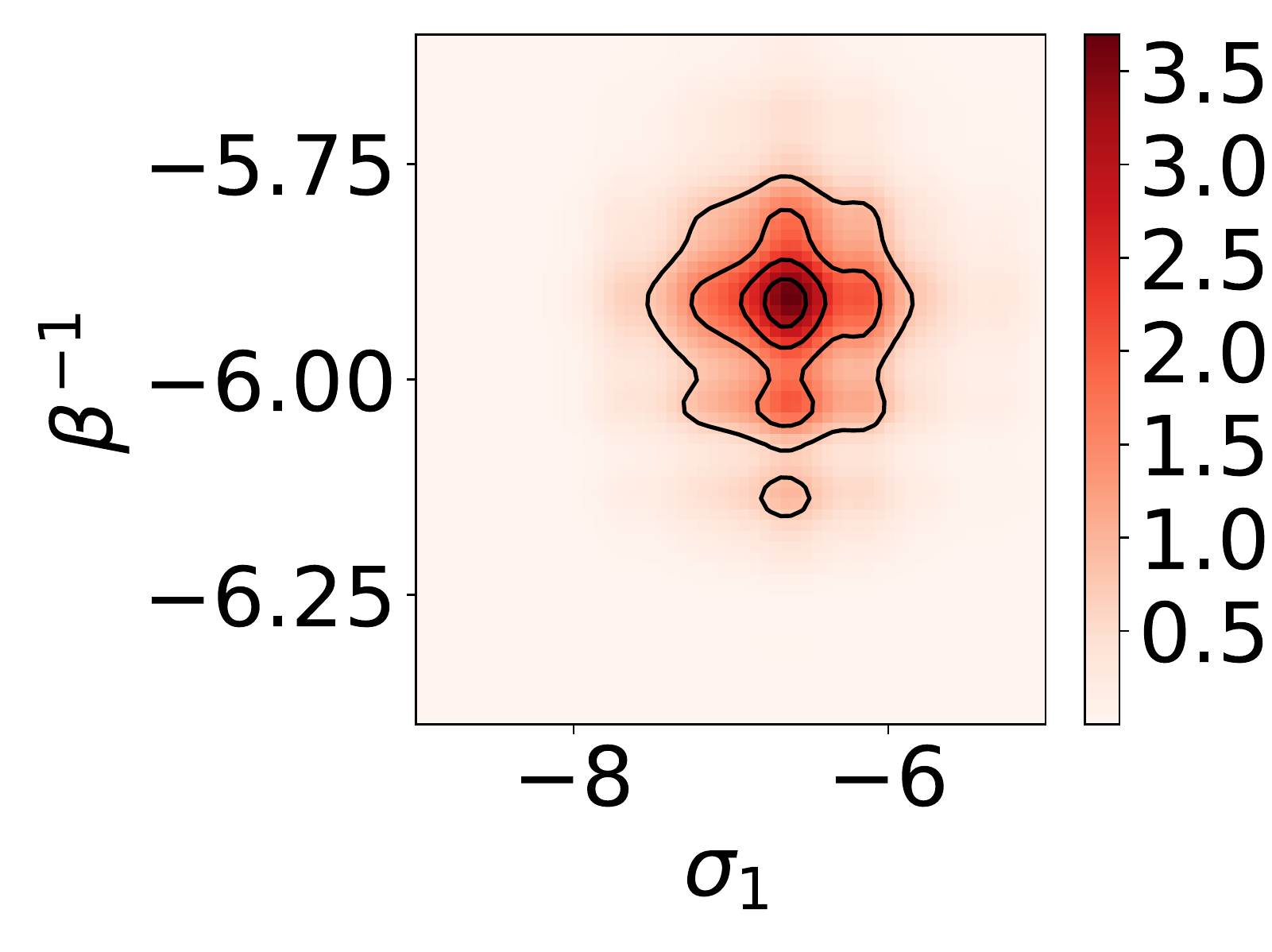}
	}
	\subfigure[Case 3: $\theta_1$ v. $\theta_2$]{
		\label{fig:Toy_mix2_Joint2}
		\includegraphics[width=30mm]{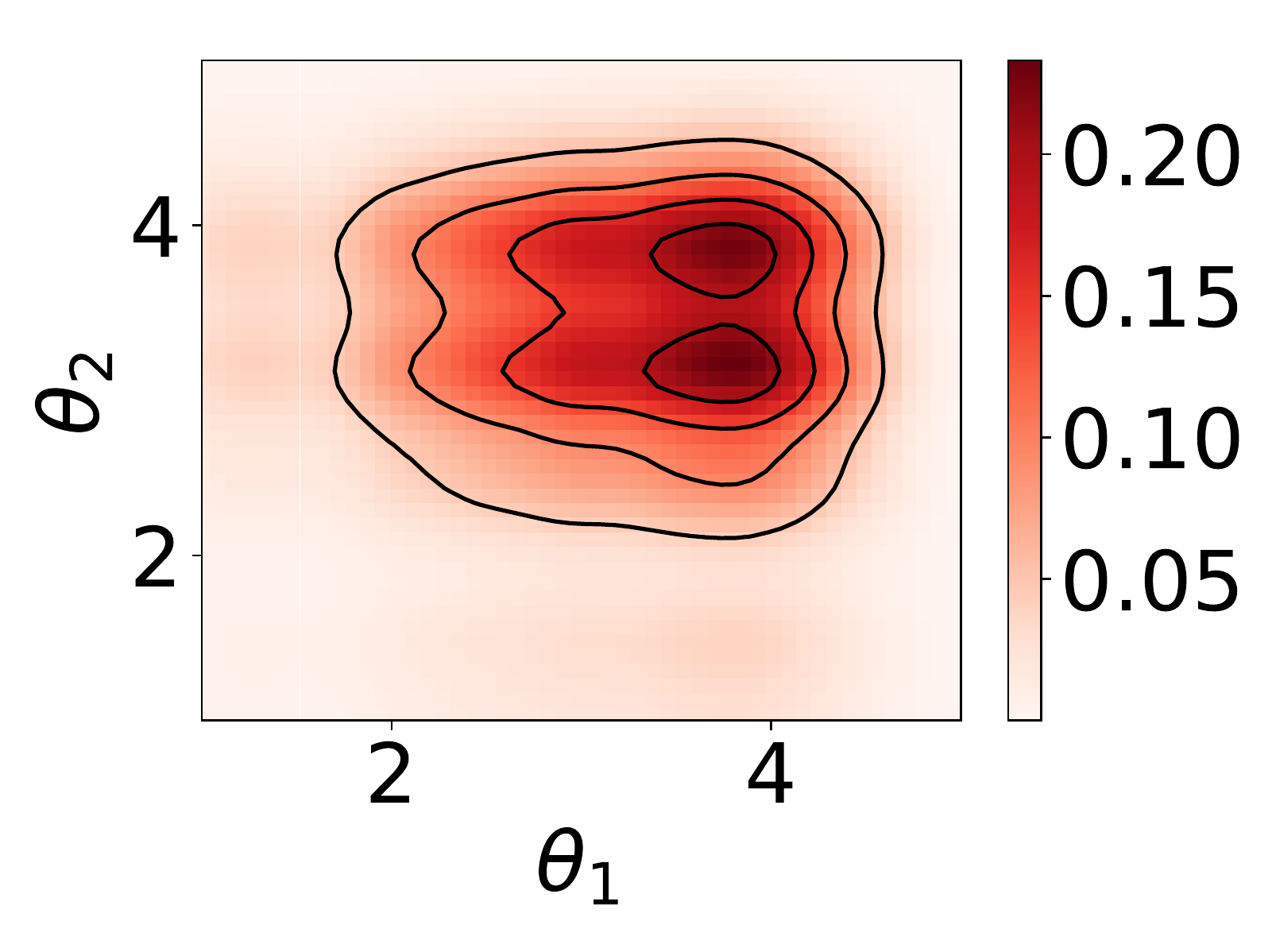}
	} 
	\subfigure[Case 3: $\theta_S$ v. $\beta^{-1}$]{
		\label{fig:Toy_mix2_Joint3}
		\includegraphics[width=30mm]{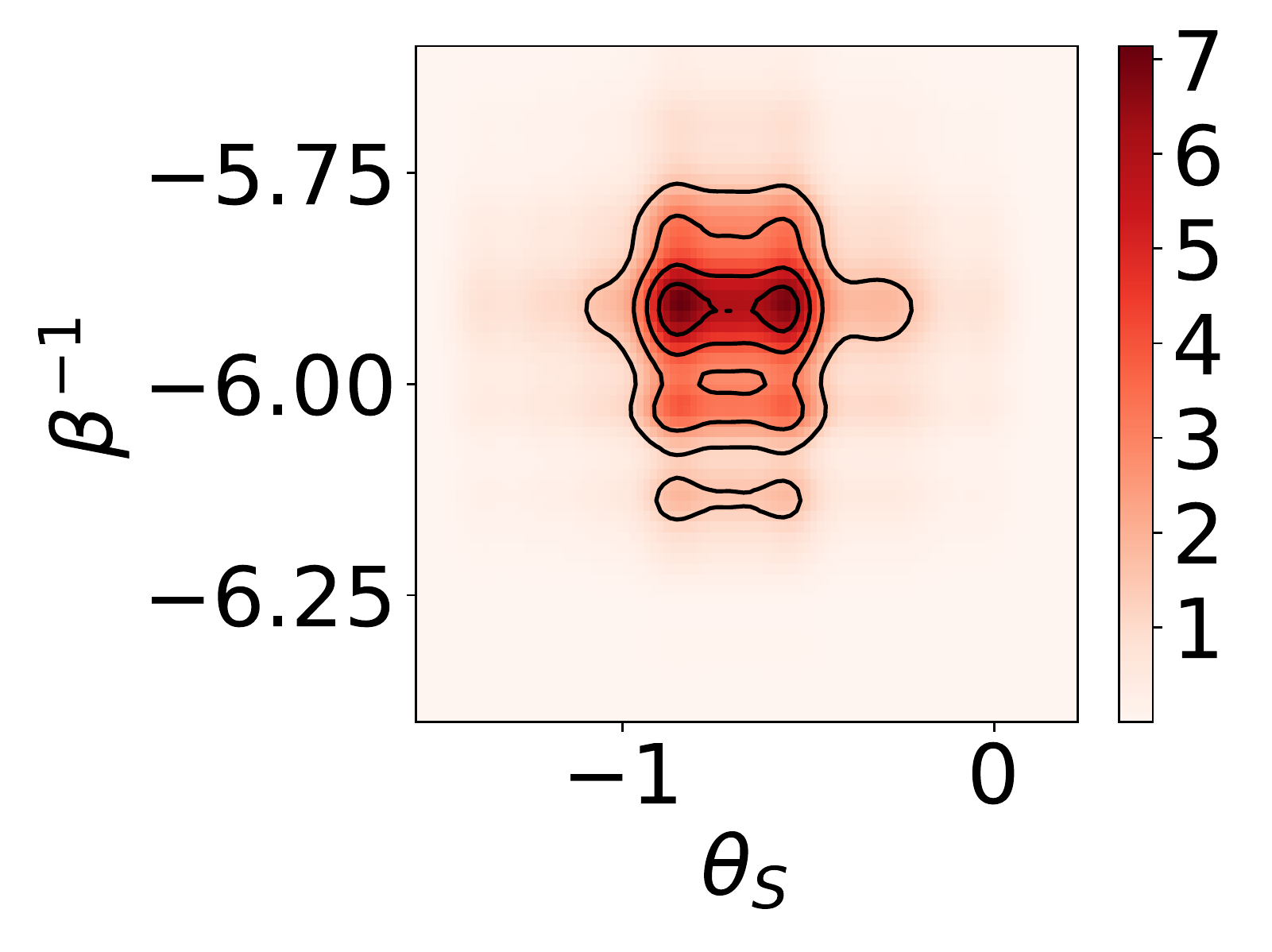}
	}
	\caption{The hyperparameter joint posterior distributions for different pairs. The three rows correspond to the three data generating cases.} 
	\label{fig:Toy_posteriors}
\end{figure*}

We generate a supervised training set using known functions with artificially added noise. We choose a set of $N=30$ input points $\bx_n \in \mathbb{R}$, linearly spaced between $0$ and $4\pi$, and use the data generating function:
\begin{equation}
f_{n,i} = \left\{ 
\begin{array}{ll}
\zeta_i\cos(F_i\bx_n) \qquad\qquad \hfill \text{ if } i=1,2,3\\
\zeta_i\sin(F_i\bx_n) \qquad\qquad \hfill \text{ if } i=4,5,6
\end{array},
\right.
\label{eq:Toy1_datagen.}
\end{equation}
with the coefficients sampled as $\zeta_i\sim\mathcal{U}\left(0,1\right)$. We then obtain  noise corrupted responses $y_{n,i} = f_{n,i}+\epsilon_{n,i}$, where $\epsilon_{n,i}\iidsim\mathcal{N}\left(0,0.05^2\right)$.

We choose to use squared exponential kernels to measure correlations in the input and latent spaces, with the addition of white noise (for numerical stability) on the preceding:
\begin{align*}
\begin{split}
k_z\left(\bx,\bx';\bsigma\right)=&\exp\left(-\frac{1}{2}\sum_{l=1}^{\kx}\sigma_{l}(x_{l}-x'_{l})^2\right)  + \varepsilon \delta\left(\bx,\bx'\right), \\
k_f(\bz,\bz';\btheta)=&\theta_{S} \exp\left(-\frac{1}{2}\sum_{j=1}^{\kz}\theta_{j}(z_j-z'_j)^2\right).
\end{split}
\label{eq:Toy_well_specified_kernels.}
\end{align*}
For identifiability, we fix the magnitude of kernel on the input space to one. Throughout we set the latent dimension $\kz=2$ for each trigonometric function in Eq. \eqref{eq:Toy1_datagen.}. We learn these functions under multiple parameterisations:
\begin{enumerate}
	\item \textbf{Case 1} - A well-specified example where $F_i = 1  \forall i$.
	\item \textbf{Case 2} - A poorly-specified example where $F_i\sim\mathcal{U}\left(0.8,1.2\right)  \forall i$.
	\item \textbf{Case 3} - A poorly-specified example where $F_i \sim \mathcal{U}\left(0.7,1.3\right)  \forall i$.
\end{enumerate} 


\begin{figure*}
	\centering
	\subfigure[True]{
		\label{fig:Toy_mix2_True}
		\includegraphics[width=35mm]{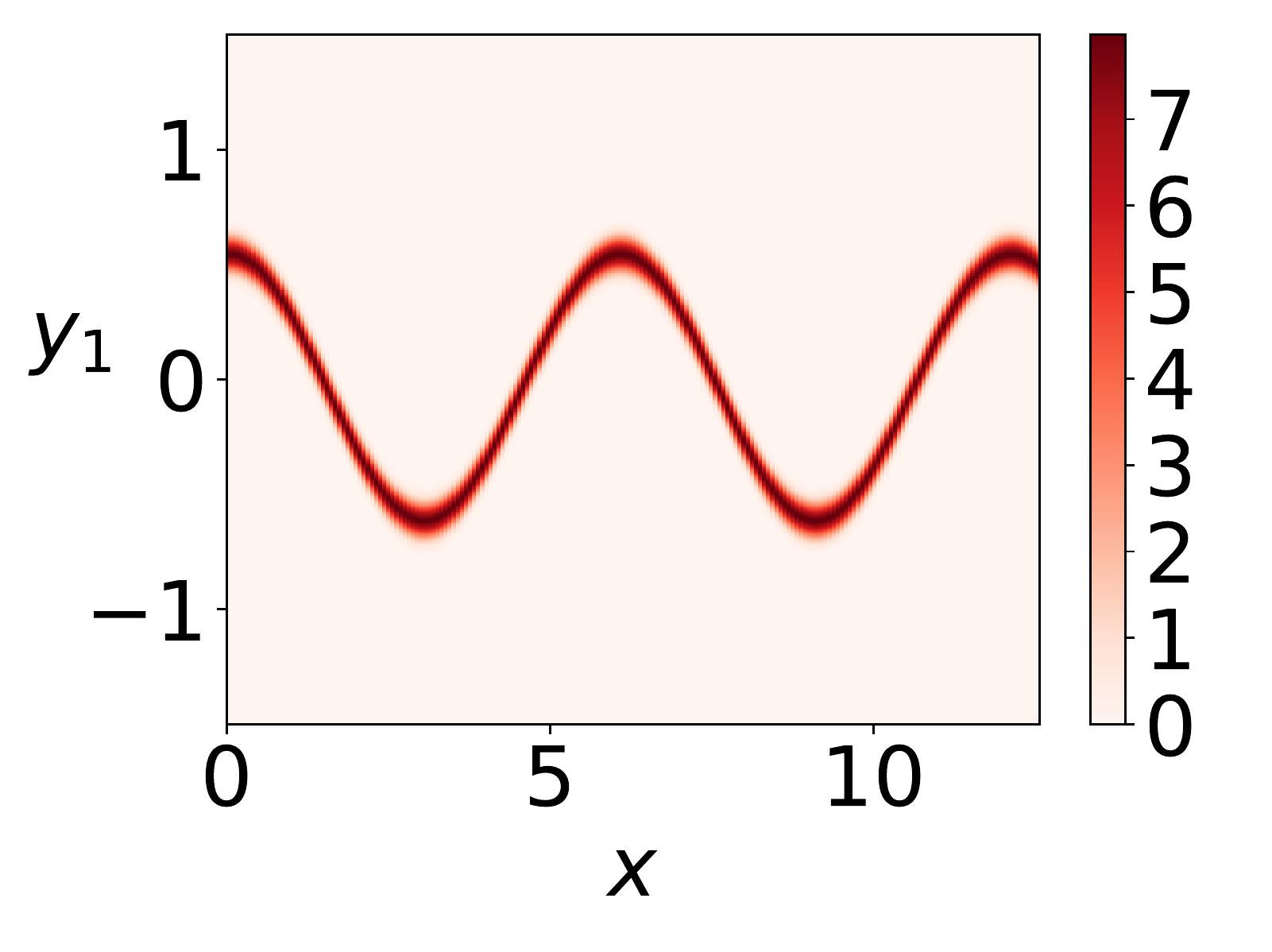}
	}
	\subfigure[Marginalized]{
		\label{fig:Toy_mix2_Marginal}
		\includegraphics[width=35mm]{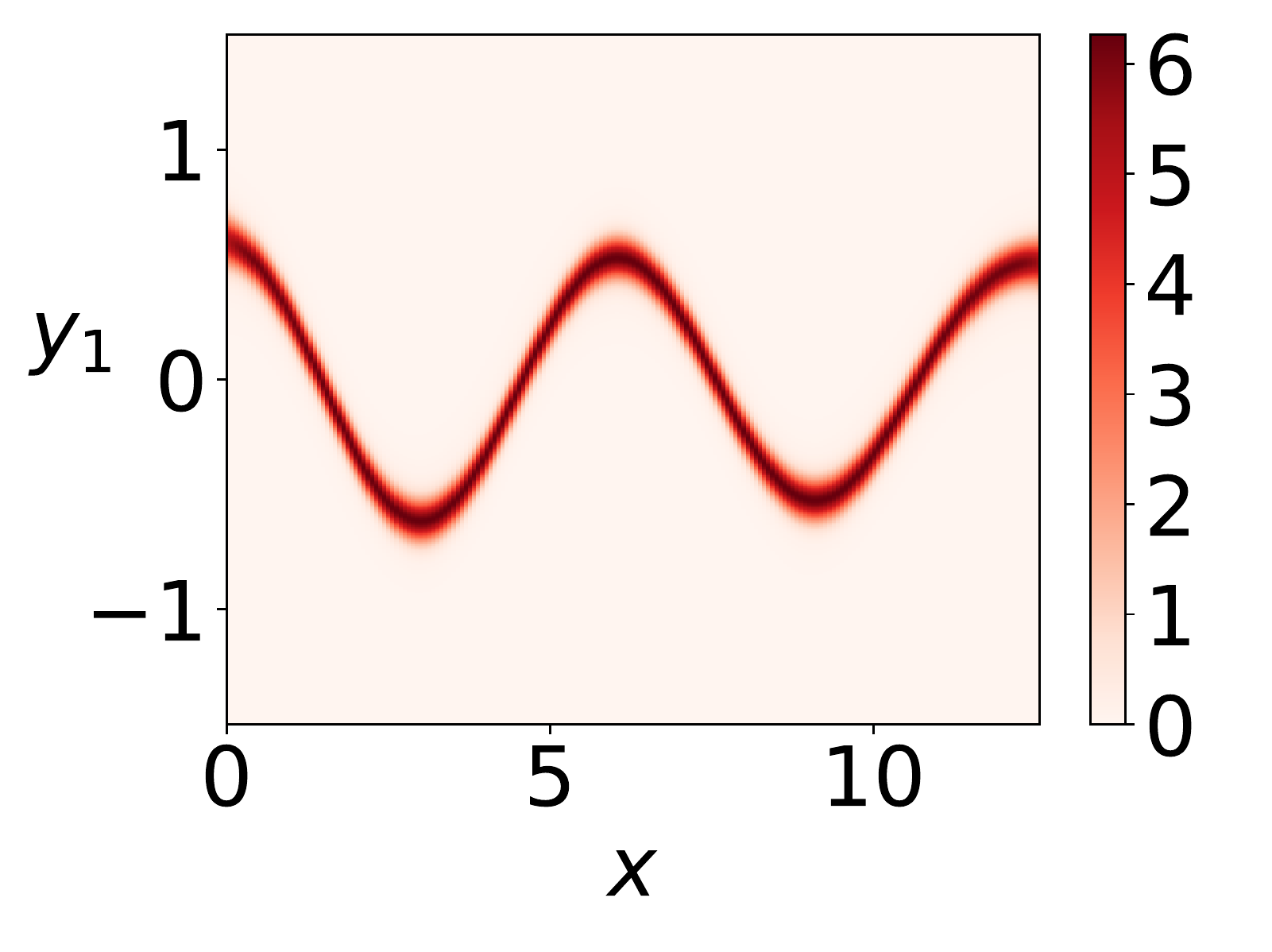}
	} 
	\subfigure[Maximum Likelihood]{
	    \label{fig:Toy_mix2_ML}
		\includegraphics[width=35mm]{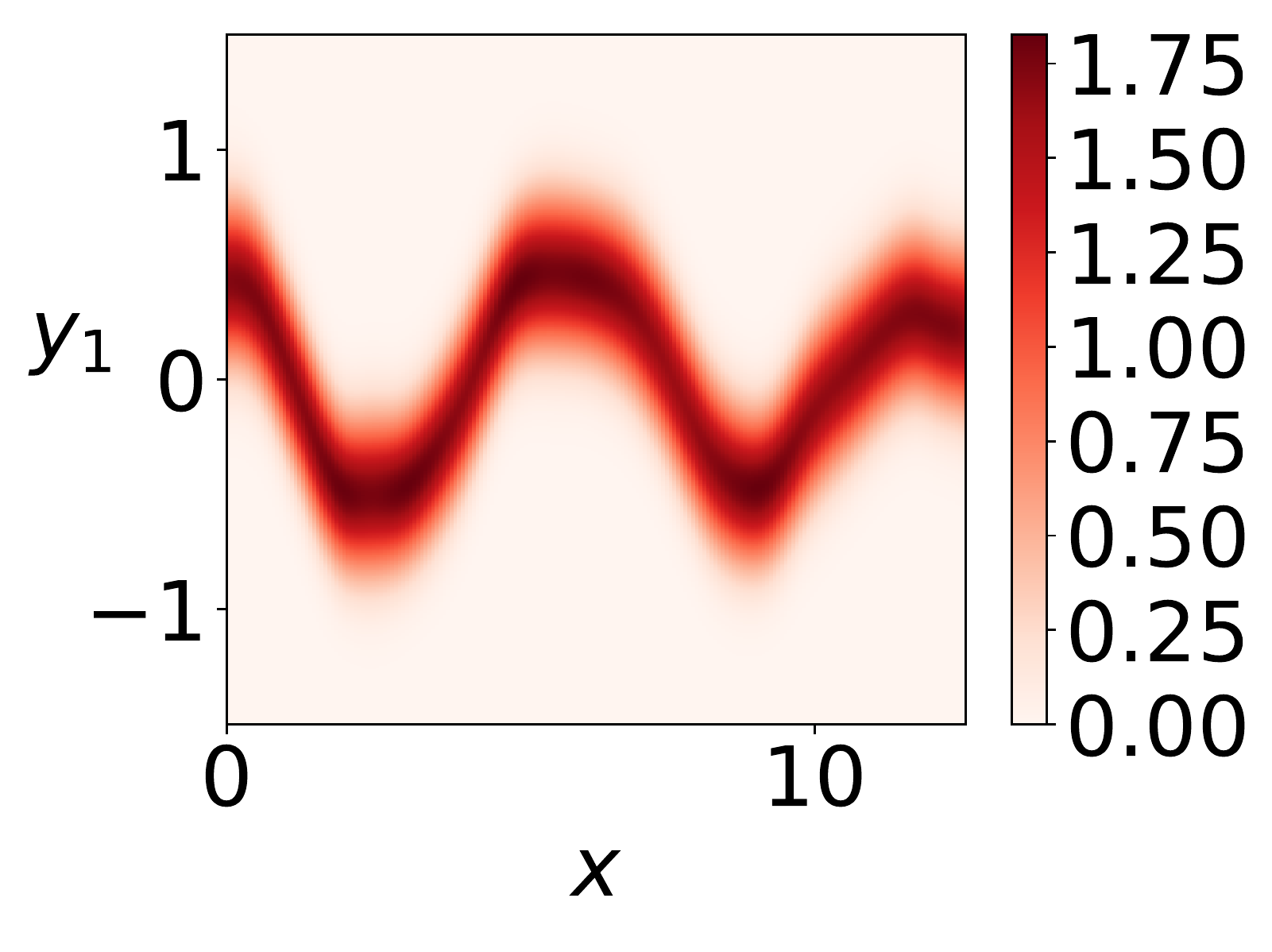}
	}
	\caption{A heat map of the true predictive density of the first feature, the estimated predictive density obtained from the proposed marginalised approach, and the variational approach with maximum likelihood estimates, for the third case.}
	\label{fig:Toy_mix2_pred1}
\end{figure*}

In the first case, we keep $F_i$ constant, showing the improvements in predictions of the fully Bayesian inference scheme that marginalises over the hyperparameters, even in this well-specified example where the variational scheme with maximum likelihood estimates of the hyperparameters should be able to make adequate predictions. We then show two cases where $F_i$ is sampled from increasing uniform intervals where we expect point estimates of the hyperparameters to give an inadequate predictive distribution. This demonstrates the ability to capture multi-modal posteriors with the proposed inference scheme, whilst also demonstrating the improved uncertainty quantification in predictions with marginalisation using our methodology over maximum likelihood. In all cases we used a Gamma prior on all hyperparameters, with a suitably large support, and a log transformation in the random walk proposals. We repeated the experiment for a number of prior parameterisations and found prediction accuracy was not sensitive to prior choice.


We choose $g_0 = 200$, $n_0 = 500$ and run four chains in parallel for $1000$ iterations each, discarding the first $250$ as a burn-in. We start each chain at the ML parameters with noise, and explore the posterior distribution using two Gibbs blocks, $\pmb{\xi}_1=\left(\sigma_1, \theta_1,\hdots,\theta_{\kz}\right)^T$ and $\pmb{\xi}_2= \left(\theta_S,\beta\right)$, using Algorithm~\ref{Alg:AMAlgorithm.}.  Rather than re-optimize the variational distribution once per cycle, we re-optimize after each full conditional sample. Trace and autocorrelation plots (not shown) demonstrated good mixing. If necessary, mixing could be further improved by splitting the hyperparameters into smaller blocks. However, this comes at the price of an increased computational cost.

The joint posterior distribution for different pairs of hyperparameters is depicted in \fref{Toy_posteriors}, where the rows correspond to the three cases.
Specifically, the pairs include the input lengthscale and model noise ($\sigma_1,\beta^{-1}$); latent lengthscales ($\theta_1,\theta_2$); and the signal variance and model noise ($\theta_S, \beta^{-1}$). When the maximum likelihood value lies in the scale, we mark it with a cross. Notice the multi-modal posteriors for the poorly-specified cases.

The marginalized predictive densities are obtained from the methodology detailed in \sref{Predictions}. \fref{Toy_mix2_pred1} displays a heat map of the true predictive density for the first feature along with the estimated predictive density through marginalisation and with maximum likelihood estimates for the third case; the variational approach with maximum likelihood estimates clearly overestimates the uncertainty. 
In general, we note a marked increase in accuracy of the predictive density with marginalisation across all features and particularly for the poorly specified examples, which is summarised in \tref{Toy_error},
 where we use the normalised mean squared error across each binned density. The table elements give a pair of percentage errors, where the first pertains to our inference scheme whilst the second to a maximum likelihood prediction. We note this optimisation is non-convex and may not reflect a global optimum despite multiple initialisations, but argue that this further necessitates posterior sampling.

\subsection{Case study: biomarkers of depression during pregnancy}
\label{sec:SingleCell.}

Depressive symptoms during pregnancy and postpartum are routinely investigated by the Edinburgh Postnatal Depression Scale (EPDS), a 10 item questionnaire. The American Academy of Pediatrics recommends an EPDS score of 10 or above as a cut off value to identify a positive screen of perinatal depression. The reported performance characteristics at this cut-off point are: sensitivity of 0.81 and specificity of 0.86 for detection of both major and minor depression and a sensitivity of 0.917 and specificity of 0.77 for major depression only \cite{cox1987detection}. There is an urgent need for better characterization of the disease process and development of novel biomarkers for use in routine diagnostics suitable for early identification of patients.

\begin{table*}
	\centering
	\begin{tabular}{||c|| c | c | c ||} 
		\hline
		& $F_i=1$ &  $F_i\sim \mathcal{U}\left(0.8,1.2\right)$  & $F_i\sim \mathcal{U}\left(0.7,1.3\right)$ \\ [0.5ex] 
		\hline\hline
		Feature 1 & (14.543\%, 28.38\%)  & (29.014\%, 64.603\%) & (31.078\%, 100.17\%)  \\ 		\hline
		Feature 2 & (12.136\%, 20.966\%) & (17.182\%, 72.581\%) & (19.235\%, 106.04\% )  \\		\hline
		Feature 3 & (14.529\%, 31.03\%)  & (12.116\%, 89.269\%) & (12.856\%, 94.032\%) \\		\hline
		Feature 4 & (10.067\%, 26.164\%) & (15.159\%, 60.252\%) & (16.915\%, 85.664\%)   \\		\hline
		Feature 5 & (5.9431\%, 18.077\%) & (10.227\%, 86.005\%) & (12.358\%, 94.287\%)   \\ 		\hline
		Feature 6 & (11.968\%, 27.708\%) & (12.441\%, 64.447\%) & (14.107\%, 101.1\%)   \\ [0.5ex]		\hline
	\end{tabular}
	\label{tab:Toy_error}
	\caption{Percentage error pairs. Each column corresponds to a different example case. The first element of each pair corresponds to the error for the marginalised predictive density, whereas the second corresponds to the maximum likelihood predictive density.}
\end{table*}

A preliminary dataset of 77 pregnant women was collected containing antenatal EPDS scores (obtained at 24-28 weeks of pregnancy) and a set of circulating hormonal signals of pathogenic processes associated with depression as well as genetic variants exhibiting single nucleotide polymorphisms (SNPs) of genes previously associated with depression. The aim is to identify relevant biomarkers and SNPs associated to perinatal depression to be collected in larger future studies. Specifically, the response of interest is the antenatal EPDS score, and biomarkers include: IL-6 (pg/ml), allopregnanalone (ng/ml) and its precursor progesterone (ng/ml), leptin (ng/ml) and brain derived neurotrophic factor (BDNF) (pg/ml), while SNPs include: corticotropin releasing hormone receptor type 1 (CRH-R1) rs242924 (G$>$T) and rs173365 (A$>$G); serotonin transporter (SLC6A4) rs6354 (G$>$T), rs2020936 (G$>$A), 5-HTTLPR (serotonin-transporter-linked polymorphic region) S-allele; glucocorticoid receptor (NR3C1) bcl1 (rs41423247, C$>$G).
 We set SNPs to binary values for wild type and non wild type genetic variations. 
 For this preliminary study, we anonymize these covariates, referring to them by their associated, unlabelled length scales $\sigma_i$. Data is normalised to the unit interval to ensure inputs are comparable. 

We first motivate the use of the supervised GPLVM by benchmarking against other standard methods. The mean square errors in the EPDS scores averaged over $500$ cross validation random subsets consisting of $5$ test points were: $37.3251$ for a GP with ML estimates; $31.9501$ for a random tree; and $30.0016$ for a supervised GPLVM with ML estimates. 

The same kernels are used in this example, and we use Gamma priors, specifically  $Ga\left(2,10\right)$ for all length scales; $Ga\left(2, 0.33\right)$ for $\theta_S$; and $Ga\left(3,0.02\right)$ for $\beta$.
We ran four chains for $3000$ iterations each. Selected summary statistics for the hyperparameter posterior distributions are given in \tref{Sum}. 
Based on the ML estimates, one would conclude that $\sigma_3$ is a significant predictor for antenatal depression, while the other inputs show low relevance. However, the posterior analysis reveals that there is significant uncertainty in every length scale, and contrary to ML analysis, we cannot identify a significant subset of inputs due to the large uncertainty.
In addition, we also observed similar uncertainties using empirical priors, where the prior mean was chosen to be equal to the ML estimates for a random choice of seed.

\begin{table}[h!]
  \centering
	\begin{tabular}[b]{||c||HHHc|c|Hc||}\hline
		& $a$ &  $b$  & Prior Mean & ML & Mean & Standard deviation & 95\% credible interval\\ \hline \hline
		$\sigma_1$ 	 & 3 & 0.0011 & 0.0034 & 0.0034 & 21.1994 & 14.8377 & (3.241,57.797) \\ \hline
		$\sigma_2$ 	 & 3 & 0.063  & 0.1896 & 0.1907 & 19.1194 & 13.8179 & (2.350,54.803)\\ \hline
		$\sigma_3$   & 3 & 7.491  & 22.4754 & 22.2445 & 20.9894 & 13.4691 & (3.488,54.395) \\ \hline
		$\sigma_4$ 	 & 3 & 0.00387& 0.0116 & 0.0117 & 20.9593 & 13.5434 & (3.759,56.413)\\ \hline
		$\sigma_5$   & 3 & 0.0017 & 0.0050 & 0.0051 & 21.2538 & 13.8772 & (3.876,58.183)\\ \hline
		$\sigma_6$   & 3 & 0.0005 & 0.0015 & 0.0015 & 21.5469 & 13.9270 & (3.060,54.754)\\ \hline
		$\sigma_7$ 	 & 3 & 0.00047& 0.0014 & 0.0014 & 18.0555 & 13.1112 & (1.597,51.817)\\ \hline
		$\sigma_8$   & 3 & 0.00037& 0.0011 & 0.0011 & 18.9564 & 12.8047 & (2.990,51.713) \\ \hline
		$\sigma_9$   & 3 & 0.00097& 0.0029 & 0.0030 & 21.8519 & 13.8486 & (3.880,56.888) \\ \hline
		$\sigma_{10}$& 3 & 0.0209 & 0.0627 & 0.0624 & 18.0718 & 13.1209 & (2.623,53.283)\\ \hline
		$\sigma_{11}$& 3 & 0.00067& 0.0020 & 0.0020 & 21.7863 & 15.3328 & (3.162,61.481)\\ \hline
		$\theta_1$   & 3 & 1/3    & 1.000  & 1.2858 & 20.7325 & 15.6269 & (1.832,55.323)\\ \hline
		$\theta_s$   &   &        &        & 0.2173 & 0.5313  & 0.4306 &  (0.079,1.624)\\ \hline
		$\beta^{-1}$ & 4 & 1/400  & 0.0100 & 0.0062 & 0.0593  & 0.0096 & (0.043,0.080)\\ [0.5ex]
		\hline
	\end{tabular}
	\caption{Summary statistics of the posterior distributions with ML estimates for comparison.}
	\label{tab:Sum}
\end{table}


%
%
\section{Discussion and concluding remarks}
\label{sec:Conclusion.}

In models with strong correlations between parameters, Gibbs sampling is known to perform poorly \cite{lawrence2009efficient}. Strong correlations between variables can result in inefficient mixing and slow convergence, and  dependence in hierarchical models can lead to local behaviour of the tuning parameters, which cannot be adapted without breaking detailed balance. Through the use of a pseudo-marginal scheme, the high correlations between latent variables and hyperparameters are broken.  

In recent years, deep learning has become a popular area of research. Many deep learning models, such as deep Gaussian processes, rely on variational approximations, both for scaling to large data sets and for analytic tractability. Although the methodology proposed here should readily extend to many such models, when the parameter space is of high dimension, we would suggest the  use of a pseudo Hamiltonian Monte Carlo scheme~\cite{lindsten2016pseudo}.

By employing  the KL divergence in the variational approximation, we  underestimate the latent variable posterior variance. This does not affect the pseudo-marginal algorithm, which has MC convergence guarantees. Of course, the closer the approximation is to the marginal, the faster the chain converges. Similarly our predictions are unaffected as, after taking advantage of pseudo-marginalisation to collapse our Gibbs sampler, we sample the latent variables using elliptical slice sampling.

Our scheme comes at the cost of an additional computational burden, increasing computational time from an order of minutes to hours. The cost of our variational approximation can be reduced by using stochastic gradients, fewer VB optimisation iterations, or performing the variational approximation less frequently. Whilst these changes will slow the convergence of our Markov Chain, we had no problems in our examples. Additionally, we can easily parallelise our MCMC scheme, leading to a significant decrease in computational time.  

We presented simulated examples which demonstrate the significant improvements which can be obtained through our scheme, 
particularly in the poorly-specified examples when point estimates of hyperparameters are insufficient.  Additionally, we presented a case study for predicting Edinburgh Postnatal Depression Scale (EPDS) scores given a set of circulating hormonal signals of the pathogenic processes and genetic variants exhibiting single nucleotide polymorphisms. We found that whilst ML estimates strongly suggested a significant input, our scheme demonstrated that the uncertainty in these estimates is too great to reach this conclusion.


\bibliography{references}
\bibliographystyle{icml2018}

\end{document}